%% file: main.tex
\documentclass[10pt,twocolumn,letterpaper]{article}
\usepackage{cvpr}              

\input{preamble}

\definecolor{cvprblue}{rgb}{0.21,0.49,0.74}
\title{Rethinking Diffusion Model-Based Video Super-Resolution: Leveraging Dense Guidance from Aligned Features}
\author{Jingyi Xu$^{12}$\thanks{\noindent Equal contribution. {\dag}Corresponding author. \\ \indent Work was done during Jingyi Xu’s internship at Alibaba.  \\
\indent Work is supported by  National Natural Science Foundation of China under Grants 62522101, 62372024 and 62231002, and Beijing Nova Program.}\quad
Meisong Zheng$^{2}$\footnotemark[1]\quad
Ying Chen$^{2\dagger}$\quad
Minglang Qiao$^{1}$ \quad
Xin Deng$^{13\dagger}$ \quad
Mai Xu$^{1}$ \\
$^{1}$Beihang\thinspace University $^{2}$Alibaba\thinspace Group\thinspace -Taobao \thinspace\&\thinspace Tmall\thinspace Group \\$^{3}$State\thinspace Key\thinspace Laboratory\thinspace of\thinspace Virtual\thinspace Reality\thinspace Technology\thinspace and\thinspace Systems\\
{\tt\small \{jingyixu,\!minglangqiao,\!cindydeng,\!MaiXu\}@buaa.edu.cn,\{meisongzheng,\!yingchen\}@alibaba-inc.com}
}
\begin{document}
\maketitle
\input{sec/0_abstract}    
\input{sec/1_intro}
\input{sec/2_related}
\input{sec/3_finding}

\input{sec/4_method}
\input{sec/5_experiment}
\input{sec/6_conclusion}
{
    \small
    \bibliographystyle{ieeenat_fullname}
    \bibliography{main}
}
\end{document}

%% file: preamble.tex
\usepackage{ulem}
\usepackage[utf8]{inputenc} 
\usepackage[T1]{fontenc}    
\usepackage{hyperref}       
\usepackage{url}            
\usepackage{booktabs}       
\usepackage{amsfonts}       
\usepackage{nicefrac}       
\usepackage{microtype}      
\usepackage{xcolor}         
\usepackage{graphicx}
\usepackage{amsmath} 
\usepackage{bm}
\usepackage{amsthm,amsmath,amssymb}
\usepackage{mathrsfs}
\usepackage[table]{xcolor}
\definecolor{mygray}{gray}{.9}
\usepackage{multirow}
\usepackage{graphicx,wrapfig}

\usepackage{xcolor}



%% file: sec/0_abstract.tex
\begin{abstract}
Diffusion model (DM) based Video Super-Resolution (VSR) approaches achieve impressive perceptual quality. However, existing DM-based VSR methods over-prioritize perceptual synthesis while neglecting fidelity gains from accurate alignment and sufficient compensation.
In this paper, within the DM-based VSR pipeline, we revisit the role of alignment and compensation between adjacent video frames and reveal two crucial observations: (a) the feature domain is better suited than the pixel domain for information compensation due to its stronger spatial and temporal correlations, and (b) warping at an upscaled resolution better preserves high-frequency information, but this benefit is not necessarily monotonic.
Therefore, we propose a novel Densely Guided diffusion model with Aligned Features for Video Super-Resolution (DGAF-VSR), with an Optical Guided Warping Module (OGWM) to maintain high-frequency details in the aligned features and a Feature-wise Temporal Condition Module (FTCM) to deliver dense guidance in the feature domain. Extensive experiments on synthetic and real-world datasets demonstrate that DGAF-VSR surpasses state-of-the-art methods in key aspects of VSR, including perceptual quality (35.82\% DISTS reduction), fidelity (0.20 dB PSNR gain), and temporal consistency (30.37\% tLPIPS reduction). The source codes are available at \url{https://github.com/tszssong/DGAF-VSR}.
\end{abstract}

%% file: sec/1_intro.tex
\vspace{-2em}
\section{Introduction}
\label{sec:intro}

Video Super-Resolution (VSR) has gained significant attention due to the increasing demand for high-quality videos.
Unlike single-image super-resolution (SISR) techniques, VSR leverages temporal information from adjacent frames, which are highly correlated but spatially misaligned, to enhance both spatial detail recovery and temporal consistency.
Traditional VSR approaches predominantly rely on handcrafted alignment and fusion mechanisms to compensate for inter-frame motion 
\cite{baker1999super,fransens2007optical,brox2004high,liu2013bayesian}. 
However, these methods suffer from limited feature extraction and alignment capabilities, inevitably leading to high-frequency information loss and reconstruction artifacts. 
To overcome these limitations, deep learning-based VSR methods have advanced rapidly in recent years \cite{chan2021basicvsr,liang2022rvrt,tian2020tdan,wang2019edvr}. 
These approaches focus on reconstructing high-frequency details of each frame by leveraging adjacent frames through various deep neural networks. Specifically, Chan \textit{et al.} \cite{chan2021basicvsr} propose a streamlined VSR pipeline that incorporates bidirectional propagation, optical flow-based alignment, feature aggregation, and upsampling operations. This framework has been widely adopted in subsequent VSR networks \cite{chan2022basicvsr++,chan2022investigating,zhou2024video}. 
Despite achieving promising fidelity, these methods often produce inferior perceptual results, such as blurring and over-smoothing.

Recent breakthroughs in generative models offer new avenues for addressing this issue. In particular, diffusion model (DM) based VSR methods \cite{zhou2023upscaleavideo,rota2024StableVSR,chen2024SATeCo} demonstrate remarkable effectiveness in generating high-resolution videos with excellent perceptual quality and maintain temporal consistency. For instance, Zhou \textit{et al.} \cite{zhou2023upscaleavideo} integrate temporal layers with flow-guided propagation in a diffusion model for maintaining temporal coherence. Rota \textit{et al.} \cite{rota2024StableVSR} introduce the temporal conditioning module into a pre-trained diffusion model to enhance video stability. 
However, these approaches enhance perceptual quality at the expense of limited fidelity. Additionally, they suffer from error accumulation due to inaccurate feature alignment.

To resolve the above challenges, in this paper, a Densely Guided diffusion model with Aligned Features for Video Super-Resolution (DGAF-VSR) is proposed. DGAF-VSR consists of a Flow prediction Module (FM) for bidirectional optical flow estimation, diffusion steps for guided feature generation, and a pre-trained VAE decoder for high-resolution (HR) frame reconstruction. Each diffusion step integrates two key modules: the Optical Guided Warping Module (OGWM) and the Feature-wise Temporal Condition Module (FTCM). Notably, both the OGWM and FTCM modules are designed based on two key observations regarding effective feature alignment and guidance: the origin of feature guidance’s effectiveness and the effect of rescaling-based warping on high-frequency feature preservation. The OGWM module enables accurate feature alignment while preserving high-frequency details for richer textures. The FTCM module facilitates fine-grained information integration from aligned adjacent features through dense feature guidance.  Due to its stronger spatial and temporal correlations, the DGAF-VSR can effectively leverage information from current and adjacent frames to ensure fidelity and high-frequency reconstruction. Comprehensive experiments on diverse synthetic and real-world datasets demonstrate that DGAF-VSR achieves state-of-the-art performance in terms of perceptual quality, fidelity, and temporal consistency.

The main contributions are summarized as follows:
\begin{itemize}
    \item 
    We quantitatively analyze the source of feature guidance’s effectiveness, and the impact of rescaling and warping on high-frequency feature information, in diffusion-based VSR pipelines. These two observations motivate the design of our OGWM and FTCM.
    \item
    OGWM effectively reduces cumulative errors through a straightforward rescaling-based warping strategy, which enables accurate feature alignment.
    \item 
    FTCM effectively bridges the gap between perceptual quality and fidelity with dense guidance, which preserves the original video information during the diffusion process.
\end{itemize}

%% file: sec/2_related.tex
\section{Related works}
\label{sec:Related}

\subsection{Video Super-Resolution Networks}
Convolutional neural networks (CNNs) have dominated VSR due to their efficient local feature extraction capability \cite{xue2019TOFlow, dai2017DCN, wang2019edvr, chan2021basicvsr,chan2022basicvsr++,tian2020tdan}. Cao \textit{et al.} \cite{cao2021vsrtransformer} pioneer the adaptation of transformer architectures for VSR through their temporal mutual attention mechanism, and subsequent methods \cite{liang2024vrt, liang2022rvrt,shi2022PSRT} further advance Cao \textit{et al.} through improved modeling of global spatial and temporal dependencies. While these approaches achieve superior fidelity, they often produce inferior perceptual results, primarily manifested as blurring and over-smoothing. To solve these problems, RealBasicVSR \cite{chan2022investigating} employs various degradation strategies to synthesize training data and utilizes perceptual loss to improve perceptual quality.

Diffusion models (DMs) have demonstrated superior versatility in image generation compared to generative adversarial networks (GANs). In the context of SISR, diffusion model (DM)-based methods \cite{saharia2021SR3, wang2024StableSR, yue2025InvSR, ResShift, wu2024OSEDiff} consistently achieve enhanced perceptual quality and photorealistic textural details. However, existing DM-based VSR methods face two critical challenges: (a) Their inherent stochasticity exacerbates temporal flickering artifacts across video frames, and (b) their training pipelines require prohibitively high computational costs for video applications. 
To mitigate these limitations while preserving the rich generative priors, existing DM-based VSR approaches typically leverage off-the-shelf diffusion models. For instance, DiffIR2VR \cite{yeh2024diffir2vr} proposes a zero-shot framework that restores videos using pre-trained image restoration diffusion models, while subsequent works \cite{rota2024StableVSR, yang2024MGLD, chen2024SATeCo, zhou2023upscaleavideo} extend text-to-image diffusion models by incorporating temporal modules and fine-tuning the added parameters. Despite these efforts, alignment inaccuracies and suboptimal feature guidance still limit their performance, as detailed in Supplementary 1.

\subsection{Alignment in VSR}  
A key distinction between VSR and SISR lies in VSR's capacity to exploit temporal information across multiple frames. However, early methods \cite{7919264,fuoli2019efficientvideosuperresolutionrecurrent,Jo_2018_DUF} often overlook the critical role of spatial alignment, which is essential for effective aggregation of highly correlated but misaligned frames or features. 
Subsequent works have attempted to address this issue through various alignment strategies \cite{xue2019TOFlow,wang2019edvr,chan2021basicvsr}. For instance, EDVR \cite{wang2019edvr} introduces deformable convolution to align features at multiple scales, while BasicVSR \cite{chan2021basicvsr} employs optical flow to assist feature alignment. Inspired by prior works, subsequent works \cite{chan2022basicvsr++, liang2022rvrt} focus on flow-based alignment. 
PSRT \cite{shi2022PSRT} shows that Vision Transformers can directly process unaligned multi-frame inputs given sufficiently large window sizes and proposes a patch alignment method to optimize the performance-computation trade-off. 
Recently, IART \cite{xu2024IART} demonstrates that inaccurate warping degrades VSR performance and addresses this by designing a coordinate network to learn the warping function. However, in the context of DM-based VSR, the effectiveness of alignment strategies remains understudied. 
To tackle this challenge, this paper proposes a simple yet effective alignment strategy based on two insightful observations within the DM-based VSR pipeline.

%% file: sec/3_finding.tex
\section{Preliminaries and Observations}
\label{sec:finding}
\subsection{Diffusion Models for VSR}
Diffusion models \cite{ho2020DDPM} use a Markov chain to convert variables from simple distributions (e.g., Gaussian) to complex distributions. In latent diffusion models (LDMs) \cite{rombach2021LDM}, this process occurs in a low-resolution latent space created by an encoder $\mathit{E}(\cdot)$, and high-resolution frames are reconstructed by a decoder $\mathcal{D}(\cdot)$. Given latent features $\bm{z}_0\sim p_{data}$, the forward diffusion process gradually adds noise to these features, obtaining the noisy features $\bm{z}_t$ at diffusion step $t$ directly from $\bm{z}_0$ following: 
\begin{equation}
\bm{z}_t = \sqrt{\hat{\alpha}_t} \bm{z}_0 + \sqrt{(1-\hat{\alpha}_t)}\bm{\epsilon},\label{con:z_t}
\end{equation}
where $t \in [1,T]$,  $\bm{\epsilon}$ is the added Gaussian noise, and $\hat{\alpha}_{t}=\prod \limits_{\tau=0}^{t} \alpha_\tau$, $\alpha_\tau$ is the noise schedule. In the reverse process, taking $\bm{z}_T$ sampled from $\mathcal{N}(0,1)$ as input, the noise component $\bm{\epsilon}_\theta(\bm{z}_t,t)$ at each diffusion step $t$ is gradually estimated and removed through an iterative Markovian process. We can sample $z_{t-1}$ as:
\begin{equation}
\bm{z}_{t-1} = \frac{1}{\sqrt{\alpha_t}}(\bm{z}_{t} - \frac{1-\alpha_{t}}{\sqrt{1-\hat{\alpha}_{t}}}\bm{\epsilon}_\theta(\bm{z}_{t}, t)) + \sigma_{t}\bm{\varepsilon}.\label{con:z_t-1}
\end{equation}
Here, $\bm{\varepsilon} \sim \mathcal{N}(0,1)$ and $\sigma_t$ represents the variance schedule. In practice, according to Eq. \ref{con:z_t}, the noise-free approximation $\tilde{\bm{z}}_{t\rightarrow 0}$ can be directly predicted by projecting the denoised features $\bm{z}_{t}$ to the initial state following:
\begin{equation}
    \tilde{\bm{z}}_{t\rightarrow 0} = \frac{1}{\sqrt{\hat{\alpha}_t}}(\bm{z}_t - \sqrt{1-\hat{\alpha}_t}\epsilon_\theta(\bm{z}_t, t)). \label{con:z_0}
\end{equation} 
The VSR problem can be articulated as follows: Given a sequence of $N$ low-resolution frames $\{\bm{x}^{i}\}_{i=1}^{N}\in\mathbb{R}^{N\times h\times w \times 3}$, the objective is to reconstruct a corresponding high-resolution sequence $\{\bm{y}^{i}\}_{i=1}^{N}\in\mathbb{R}^{N\times sh\times sw \times 3}$, where $s$ represents the rescaling factor.

\subsection{Observations within DM-based Pipeline}
\label{sec:Finding}

\begin{figure}
  \includegraphics[width=0.48\textwidth]{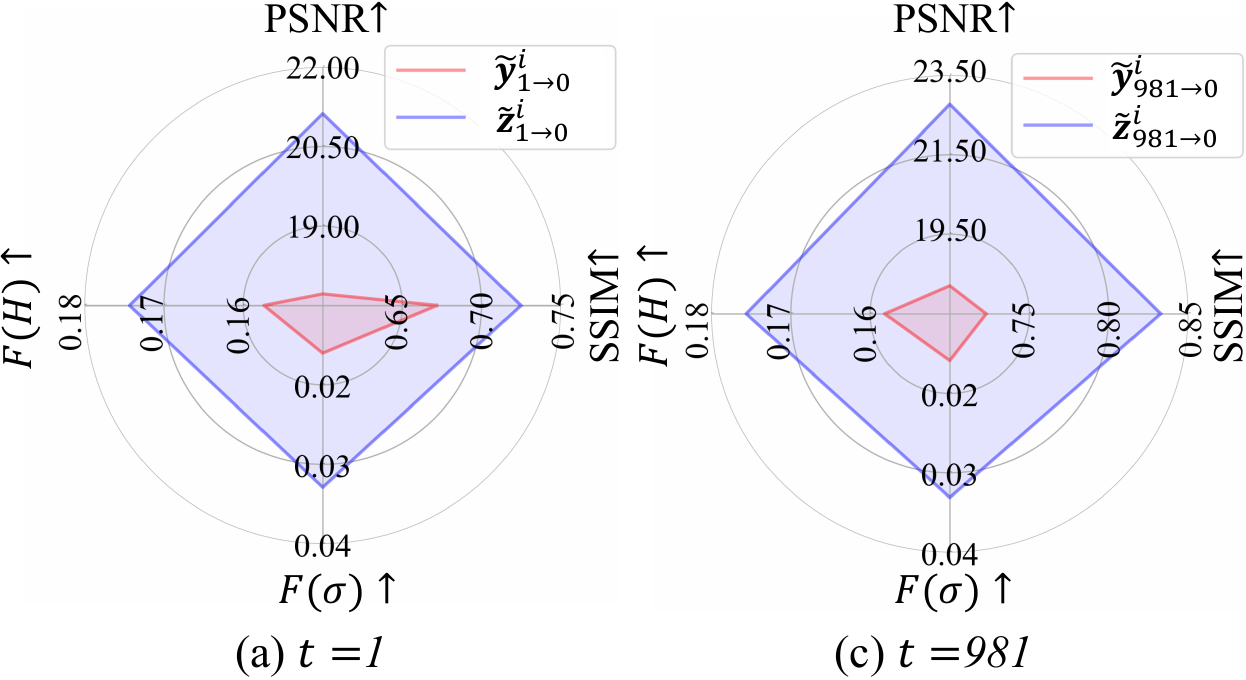}
    \vspace{-1.5em}
  \caption{Illustration of average SSIM, PSNR, $F(H)$ and $F(\sigma)$ metrics for adjacent variables across all frames in the REDS4 dataset, shown for diffusion steps (a) $t=\textit{1}$ and (b) $t=\textit{981}$. }
  \vspace{-1em}
  \label{fig:Finding1}
\end{figure}

\textbf{Observation 1: The latent feature domain outshines the pixel domain in terms of spatial and temporal correlations.} 
To verify this observation, we conduct experiments on the REDS4 dataset \cite{Nah_2019_CVPR_Workshops_REDS}. 
For the $\textit{i}$-th frame at the $\textit{t}$-th step, we employ a pre-trained DM-based VSR framework to extract two types of variables: the noise-free approximation feature $\tilde{\bm{z}}_{t\rightarrow 0}^{i}$ predicted from $\bm{z}_{t}^{i}$ and the reconstructed high-resolution frame $\tilde{\bm{y}}_{t\rightarrow 0}^{i}$. We adopt four metrics to compute the correlation between adjacent feature pairs $\tilde{\bm{z}}_{t\rightarrow 0}^{i},\tilde{\bm{z}}_{t\rightarrow 0}^{i-1}$ and frame pairs $\tilde{\bm{y}}_{t\rightarrow 0}^{i},\tilde{\bm{y}}_{t\rightarrow 0}^{i-1}$, respectively: the Structural Similarity Index Measure (SSIM) \cite{wang2004image}, the Peak Signal-to-Noise Ratio (PSNR) \cite{hore2010image}, and the Cross-Entropy ($H$) \cite{shannon1948mathematical} for pixel-level and distributional inter-variable correlation, as well as the Standard Deviation $\sigma$ \cite{altman2005standard} for intra-variable correlation. To enable consistent comparison across all metrics, we transform $H$ and $\sigma$ into $F(H)=\frac{1}{1+H}$ and $F(\sigma)=\frac{1}{1+\sigma}$. In these transformed metrics, higher values indicate stronger correlations.

Figure \ref{fig:Finding1} shows the average correlation metrics across all frames and features from diffusion steps $t=\textit{1},\textit{981}$-th diffusion steps \footnote{We select these steps to represent the early and late stages of the diffusion process.} Results across multiple video samples from REDS4 consistently show that the latent feature domain outperforms the pixel domain in all four metrics. 
For instance, at the $\textit{981}$-th diffusion step, the average PSNR between adjacent features is 25.06\% higher than that between adjacent frames. At the $\textit{1}$-th diffusion step, the $F(\sigma)$ value for features is 2.1 times that for frames. Averaged over all diffusion steps, the feature domain achieves improvements of 13.53\% in PSNR, 22.61\% in SSIM, 10.81\% in $F(H)$ and 106.50\% in $F(\sigma)$ compared to the pixel domain. These results confirm that the latent feature domain provides significantly stronger spatial and temporal correlations, making it more suitable for guidance in DM-based VSR. 

Although BasicVSR \cite{chan2021basicvsr} also advocates feature-level guidance, we presents the first quantitative validation of this principle in the context of DM-based VSR. This observation directly motivates our design of dense feature-level guidance in DGAF-VSR. For a detailed comparison with BasicVSR, see Supplementary 2.

\textbf{Observation 2: Warping at higher resolution better preserves high-frequency information, but only up to an optimal rescaling factor within the DM-based pipeline.} 
\begin{figure*}
  \includegraphics[width=\textwidth]{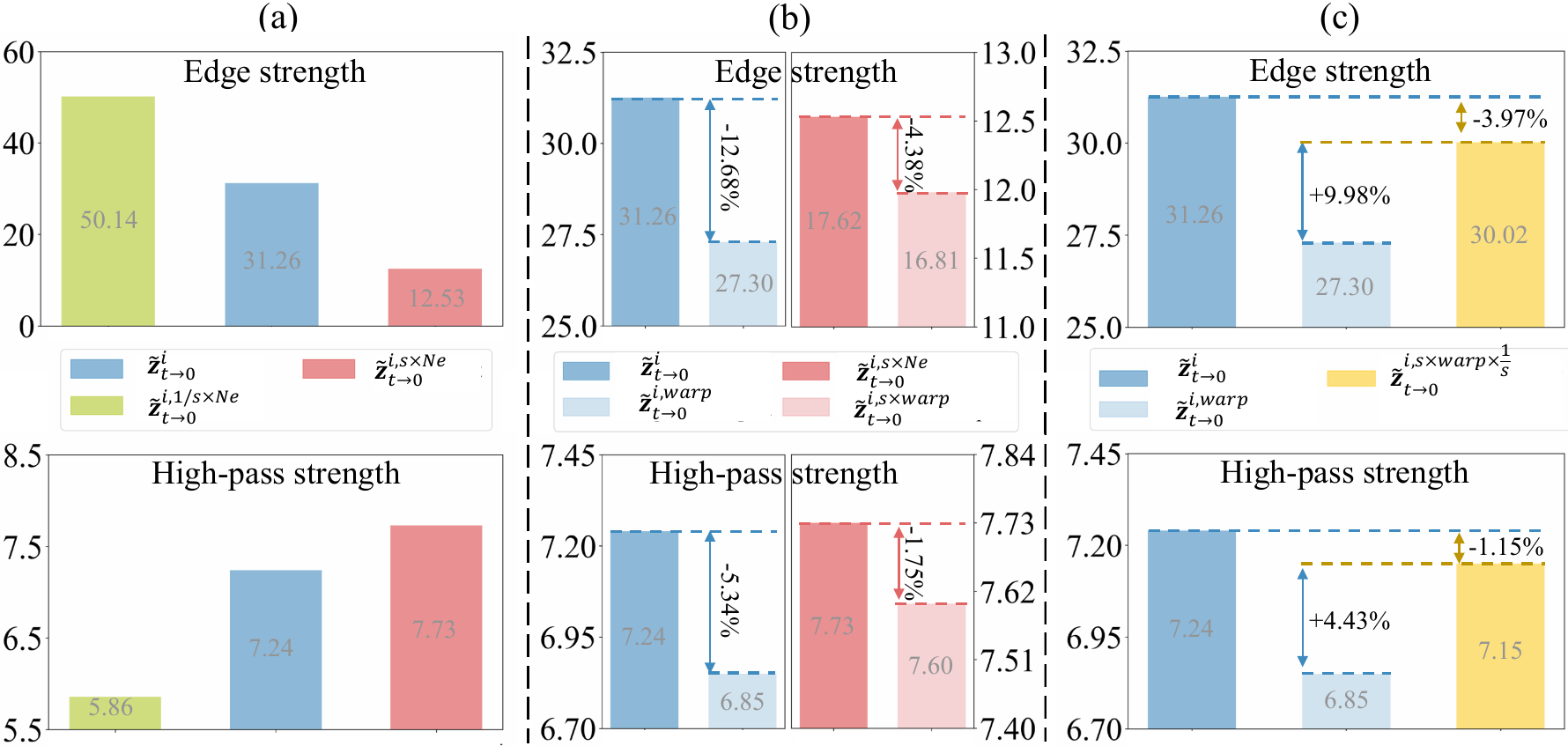}
    \vspace{-1.5em}
  \caption{Illustration of the edge strength and the high-pass strength under different alignment strategies. The impact of (a) the upscaling and downscaling operations, (b) the warping operation on features of different resolutions, and (c) the upscaling–warping–downscaling strategy.}
    \vspace{-1em}
  \label{fig:Finding2_1}
\end{figure*}
Given the fact that high-frequency information critically influences super-resolution quality \cite{zhou2018high,li2023feature,fritsche2019frequency}, we quantitatively analyze the high-frequency loss during warping. Our Observation 2 within DM-based VSR pipeline can be summarized as follows:
\begin{enumerate}
\item {Feature resolution directly affects high-frequency density.}
\item {The warping operation reduces high-frequency density within features, while high-resolution features suffer less high-frequency degradation during warping compared to low-resolution features.}
\item {The rescaling-based warping strategy effectively preserves the high-frequency information during alignment in DM-based VSR.}
\item {An optimal rescaling exists for maximizing the high-frequency preservation of the rescaling-based warping strategy.}
\end{enumerate}

To validate Observation 2, we analyze over 80,000 feature groups from REDS4 using two high-frequency metrics: edge strength (computed with Canny \cite{canny1986computational}, Sobel \cite{sobel19683x3}, and Laplacian \cite{jain1995machine} operators), and high-pass strength (computed as spectrum energy \cite{van1996modelling} after a Fast Fourier Transform (FFT) \cite{nussbaumer1982fast} followed by a high-pass filter with a threshold of 30). Higher values of both metrics indicate greater high-frequency density.

For the first point, we apply an $s\times$ upscaling and an $\frac{1}{s}\times$ downscaling operations on the original low-resolution features $\{\tilde{\bm{z}}_{t\rightarrow 0}^{i}\}_{i=1,t=1}^{N,T}$ using nearest neighbor interpolation \cite{shepard1968two} \footnote{Note that $s$ is chosen according to the fourth point.}. This process obtains two transformed feature sets: the upscaled features $\{\tilde{\bm{z}}_{t\rightarrow 0}^{i,s\times \rm{Ne}}\}_{i=1,t=1}^{N,T}$ and the downscaled features $\{\tilde{\bm{z}}_{t\rightarrow 0}^{i,\frac{1}{s}\times \rm{Ne}}\}_{i=1,t=1}^{N,T}$. The edge strength and high-pass strength are calculated for these three sets, respectively. The results are illustrated in Figure \ref{fig:Finding2_1} (a). 
As demonstrated, both metrics are sensitive to feature resolution. Therefore, all subsequent high-frequency analyses are conducted at a fixed resolution to ensure fair comparison.

For the second point, we implement bilinear warping on both $\{\tilde{\bm{z}}_{t\rightarrow 0}^{i}\}_{i=1,t=1}^{N,T}$ and $\{\tilde{\bm{z}}_{t\rightarrow 0}^{i,s\times \rm{Ne}}\}_{i=1,t=1}^{N,T}$. The resulting warped features are denoted as $\{\tilde{\bm{z}}_{t\rightarrow 0}^{i,warp}\}_{i=1,t=1}^{N,T}$ and $\{\tilde{\bm{z}}_{t\rightarrow 0}^{i,s\times warp}\}_{i=1,t=1}^{N,T}$, respectively. 
As illustrated in Figure \ref{fig:Finding2_1} (b), warping the low-resolution features causes a 12.68\% reduction in edge strength and a 5.34\% reduction in high-pass strength. In contrast, warping the high-resolution features leads to only a 4.38\% loss in edge strength and a 1.75\% loss in high-pass strength. 
These results indicate that although warping inevitably reduces high-frequency density, its impact is approximately three times milder when applied to high-resolution features than to low-resolution ones. 
This observation motivates our investigation of the third point.

For the third point, we process the rescaling-based warping features $\{\tilde{\bm{z}}_{t\rightarrow 0}^{i,s\times warp \times \frac{1}{s}}\}_{i=1,t=1}^{N,T}$ through $\frac{1}{s}\times$ nearest neighbor interpolation-based downscaling on $\{\tilde{\bm{z}}_{t\rightarrow 0}^{i,s\times warp}\}_{i=1,t=1}^{N,T}$. Figure \ref{fig:Finding2_1} (c) shows the edge strength and high-pass strength of these downscaled features. Compared to directly warping the original low-resolution features, this upscale-warp-downscale strategy improves high-pass strength by 4.43\% and edge strength by 9.98\%. 

The edge strength presented in Figure \ref{fig:Finding2_1} is computed using the Canny operator. Complete analyses using the Sobel and Laplacian operators are included in Supplementary 3.
The high-pass strength and edge strength across different diffusion steps are presented in Supplementary 4.
Details regarding the fourth point are presented in Supplementary 5.

%% file: sec/4_method.tex
\begin{figure*}
  \includegraphics[width=\textwidth]{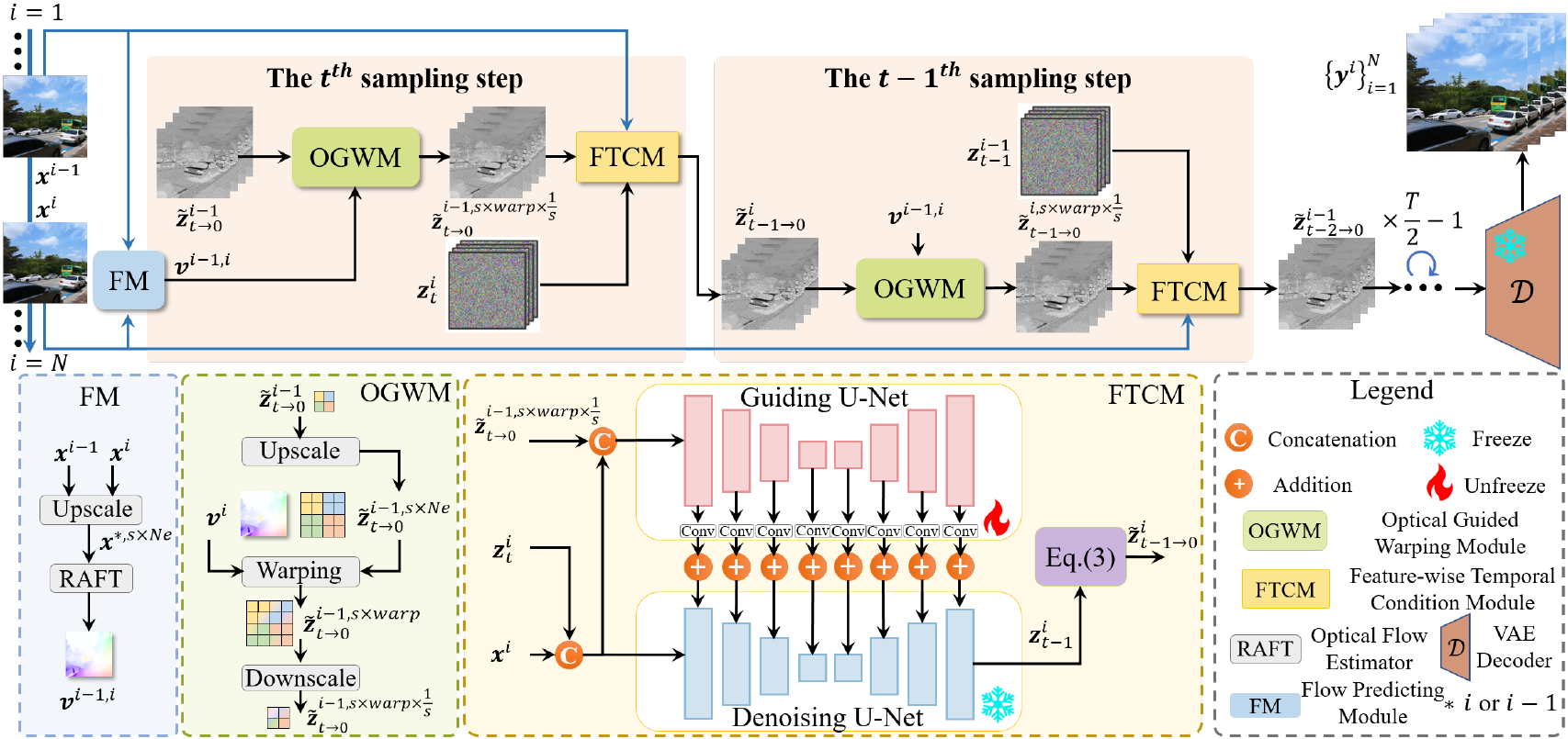}
    \vspace{-1em}
  \caption{The framework of the proposed DGAF-VSR. The network is composed of a Flow prediction Module (FM), $T$ diffusion steps, and a VAE decoder. The FM module estimates bidirectional optical flow estimation. Each diffusion step consists of one Optical Guided Warping Module (OGWM) module for feature warping process, and one Feature-wise Temporal Condition Module (FTCM) module for fine-grained information integration through dense feature guidance.}
  \vspace{-1em}
  \label{fig:methodoverview}
\end{figure*}

\section{Method}
\label{sec:method}
Based on the observations established in Section \ref{sec:Finding}, we present the Densely Guided diffusion model with Aligned Features for Video Super-Resolution (DGAF-VSR). The key ideas of our network are compared with those of MGLD-VSR \cite{yang2024MGLD} in Supplementary 6. 
As shown in Figure \ref{fig:methodoverview}, DGAF-VSR consists of a Flow prediction Module (FM), $T$ diffusion steps, and a pre-trained VAE decoder. The FM module utilizes a RAFT module \cite{teed2020raft} to calculate the compensatory optical flow $\bm{v}^{i-1,i}$ between the upscaled adjacent frames $x^{i,s\times Ne}$ and $x^{i-1, s\times Ne}$. This process effectively captures motion patterns at a higher resolution. 
The sequence of diffusion steps can be partitioned into $\frac{T}{2}$ pairs. 
Each pair includes a forward guidance process (leveraging temporal information from previous features) and a backward guidance process (leveraging temporal information from subsequent features). In each guidance process, the Optical Guided Warping Module (OGWM) is employed to align adjacent features, while the Feature-wise Temporal Condition Module (FTCM) is utilized to denoise current features under the dense guidance of aligned adjacent features. In the following, we describe the detailed architectures of these modules, using the forward guidance process as a representative example.

\subsection{Optical Guided Warping Module}
The Optical Guided Warping Module (OGWM) is built on \textbf{Observation 2}. While prior work (e.g., StableVSR \cite{rota2024StableVSR}) demonstrates that direct motion compensation in latent space often introduces artifacts, our OGWM tackles the key challenges of achieving accurate spatial alignment while preserving high-frequency information. 
As illustrated in Figure \ref{fig:methodoverview}, the OGWM executes three stages during the forward guidance process in the $t$-th diffusion step: 
\textbf{(a) Input Preparation}: The OGWM module takes $\tilde{\bm{z}}_{t\rightarrow 0}^{i-1}$ as input, which represents the noise-free approximation of the previous feature from the previous diffusion step. $\tilde{\bm{z}}_{t\rightarrow 0}^{i-1}$ is upscaled by a factor of $s=4$ using nearest neighbor interpolation, resulting in $\tilde{\bm{z}}_{t\rightarrow 0}^{i-1,s\times Ne}$. 
\textbf{(b) Feature Alignment}: The upscaled features $\tilde{\bm{z}}_{t\rightarrow 0}^{i-1,s\times Ne}$ are warped using the computed optical flow $\bm{v}^{i-1,i}$ to produce the aligned feature $\tilde{\bm{z}}_{t\rightarrow 0}^{i-1,s\times warp}$ in the latent space. 
\textbf{(c) Downscaling and Integration}: The aligned features $\tilde{\bm{z}}_{t\rightarrow 0}^{i-1,s\times warp}$ are downscaled by a factor of $s$ to obtain $\tilde{\bm{z}}_{t\rightarrow 0}^{i-1,s\times warp \times \frac{1}{s}}$. Finally, $\tilde{\bm{z}}_{t\rightarrow 0}^{i-1,s\times warp \times \frac{1}{s}}$ is fed into the FTCM module for dense feature guidance.

\begin{table*}[htbp]
\caption{Quantitative comparison on the REDS4 and Vid4 dataset for $4\times$ VSR in terms of five perceptual($\diamond$) and two fidelity metrics($\star$), with the best results in bold and second bests underlined.}
\vspace{-1em}
\label{tab:x4}
\begin{center}
\centering
\renewcommand{\arraystretch}{0.8}
  \setlength{\tabcolsep}{1.5mm}
\begin{tabular}{c|c|c|ccccc|cc}
\toprule[1.0pt]
\rowcolor{mygray} 
\multirow{-1}{*}{Dataset} &\multirow{-1}{*}{Types}
& \multirow{-1}{*}{Methods} & LPIPS$\diamond\downarrow$ & DISTS$\diamond\downarrow$ & MUSIQ$\diamond\uparrow$ &CLIP-IQA$\diamond\uparrow$ & NIQE$\diamond\downarrow$ & PSNR$\star\uparrow$ & SSIM$\star\uparrow$ \\ \hline\hline
&& Bicubic  & 0.453 & 0.186 & 26.89 & 0.304 & 6.85& 26.13 & 0.729  \\
&& EDVR \cite{wang2019edvr} &	0.178&	0.082&65.44&0.367&4.15& 31.02 & 0.879\\
&& BasicVSR \cite{chan2021basicvsr} &0.165	&0.081&	65.74	&0.371&	4.13& 31.42&	0.891		\\
&& BasicVSR++ \cite{chan2022basicvsr++} &	0.132&	0.069&	67.00 &	0.381	& 3.85& \underline{32.39}&	\underline{0.907}	\\
&\multirow{-4}{*}{\begin{tabular}[c]{@{}c@{}}Non-\\DM-\\based\end{tabular}} 
& RVRT  \cite{liang2022rvrt} &0.128	&0.067	&67.46	&0.392	&3.78 &\textbf{32.75}	&\textbf{0.911}	\\
\cline{2-10} 
&& UAV \cite{zhou2023upscaleavideo} &	0.266&	0.124	&47.94&	0.193&	4.69& 26.09&	0.724	\\
&& MGLD-VSR \cite{yang2024MGLD} &	0.141&	0.061&	67.37&	0.409&	2.81& 26.20&	0.736\\
&& StableVSR \cite{rota2024StableVSR} &	\underline{0.098}&	\underline{0.045}&	\underline{67.62}&	\underline{0.418}&	\underline{2.74}& 27.97	&0.795	\\
&& STAR \cite{xie2025star} & 0.285 & 0.117 & 66.54 & 0.266 & 3.45 & 23.01 &0.644	\\
&& DOVE \cite{chen2025dove} & 0.253 & 0.149 & 51.03 & 0.215 &4.37 & 26.01 & 0.724 \\
&& SeedVR2 \cite{wang2025seedvr2} & 0.230 & 0.099 & 64.94 & 0.284 &3.33 & 23.36 & 0.687	\\
\multirow{-11}{*}{\begin{tabular}[c]{@{}c@{}}REDS4\end{tabular}} &\multirow{-4}{*}{\begin{tabular}[c]{@{}c@{}}DM-\\based\end{tabular}} 
& Ours &	\textbf{0.095}&	\textbf{0.043}	&\textbf{67.90}	&\textbf{0.429}	&\textbf{2.66}& 28.17&	0.804\\ \hline
\toprule[1.0pt]
&& Bicubic  & 0.497 & 0.242 & 26.54 & 0.342 & 7.55 & 23.78 & 0.635 \\
&& EDVR \cite{wang2019edvr}  & 0.224  & 0.138 & 59.63 & 0.311  & 4.96 & 27.35  &  0.826 \\
&& BasicVSR \cite{chan2021basicvsr} & 0.214 & 0.136 & 60.90 & 0.320 & 4.88 & 27.24 & 0.825 \\
&& BasicVSR++ \cite{chan2022basicvsr++} & 0.189 & 0.122 & 61.50 & 0.341 & 5.04& \underline{27.79} & \underline{0.840}  \\
&\multirow{-4}{*}{\begin{tabular}[c]{@{}c@{}}Non-\\DM-\\based\end{tabular}} 
& RVRT  \cite{liang2022rvrt} &0.188	& \underline{0.114}	& 61.84	&0.374	&4.89 &\textbf{27.99}&\textbf{0.846}	\\
\cline{2-10} 
&&UAV \cite{zhou2023upscaleavideo} & 0.340 & 0.204 & 63.21 & 0.264 & 3.78 & 22.14 & 0.537 \\
&&MGLD-VSR \cite{yang2024MGLD}  & 0.243 & 0.163 & 65.07 & 0.366 & 3.20 & 23.64 & 0.641\\
&&StableVSR \cite{rota2024StableVSR}  & \underline{0.182} & 0.116 & \underline{67.20} & \underline{0.453} & \underline{3.19}& 24.47 & 0.699 \\
&& STAR \cite{xie2025star} & 0.372 & 0.171 & 59.12 & 0.314 & 4.35 & 18.70 &0.486	\\
&& DOVE \cite{chen2025dove} & 0.259 & 0.192 & 54.11 & 0.274 &6.13 & 23.07 & 0.682 \\
&& SeedVR2 \cite{wang2025seedvr2} & 0.233 & 0.128 & 64.01 & 0.383 &3.66 & 21.18 & 0.639\\
\multirow{-11}{*}{\begin{tabular}[c]{@{}c@{}}Vid4\end{tabular}}&\multirow{-4}{*}{\begin{tabular}[c]{@{}c@{}}DM-\\based\end{tabular}} 
& Ours  & \textbf{0.175} & \textbf{0.113} & \textbf{67.95} & \textbf{0.470} & \textbf{3.10}& 24.75 & 0.714 \\ \hline
\toprule[1.0pt]
\end{tabular}
\vspace{-1em}
\end{center}
\end{table*}

\subsection{Feature-wise Temporal Condition Module}
The Feature-wise Temporal Condition Module (FTCM) aggregates information from multiple frames and aligned features to provide dense motion compensation. 
Specifically, our guiding mechanism builds upon the analysis presented in \textbf{Observation 1} in Section \ref{sec:Finding} by utilizing adjacent features aligned through the OGWM module. Moreover, while existing DM-based super-resolution methods typically employ only the U-Net encoder as the guiding network \cite{rota2024StableVSR, wang2024StableSR, yang2024MGLD, he2024venhancer}, our FTCM incorporates the complete U-Net architecture, inspired by BrushNet \cite{ju2024brushnet}. The key ideas of our FTCM are compared with those of BrushNet in Supplementary 7. This design enables dense integration of adjacent information by introducing strict pixel-to-pixel constraints at different receptive fields during information extraction and reconstruction.
By leveraging the full U-Net, we ensure more effective compensation of motion information, thereby allowing our proposed network to preserve fine-grained details and produce high-quality results.

Within each diffusion step $t\in [1,T]$ of the DM model, the input low-resolution video is processed frame by frame. For instance, denoising of the $i$-th latent feature $\bm{z}^i_t$ at the $t$-th diffusion step is guided by the aligned adjacent feature $\tilde{\bm{z}}_{t\rightarrow 0}^{i-1,s\times warp \times \frac{1}{s}}$ according to the following equation: 
\begin{equation}
\bm{z}_{\!t\!-\!1\!}^{i}\!\!=\!\!DU(\!\begin{bmatrix}\bm{z}_{t}^{i},\bm{x}^{i} \end{bmatrix}\!)\!\!+\! \!Conv(\!GU(\!\begin{bmatrix}\!\bm{z}_{t}^{i},\!\bm{x}^i,\!\tilde{\bm{z}}_{t\rightarrow 0}^{i-1,s\!\times warp \!\times \!\frac{1}{s}}\!\end{bmatrix}\!)\!). \label{con:feature}
\end{equation}
$DU(\cdot)$ denotes a pretrained Denoising U-Net with frozen parameters during training, while $GU(\cdot)$ represents the Guiding U-Net with parameters that are trainable throughout the training process, as shown in Figure \ref{fig:methodoverview}. The notation $\begin{bmatrix} \bm{z}_{t}^{i} , \bm{x}^{i} \end{bmatrix}$ signifies the concatenation of the current low-resolution frame $\bm{x}^{i}$ and the current latent feature $\bm{z}_{t}^{i}$ updated from the previous diffusion step $t+\textit{1}$. Similarly, $\begin{bmatrix} \bm{z}_{t}^{i}, \bm{x}^i, \tilde{\bm{z}}_{t\rightarrow 0}^{i-1,s\times warp \times \frac{1}{s}} \end{bmatrix}$ indicates the concatenation of $\bm{z}_{t}^{i}$, $\bm{x}^i$ and the aligned adjacent feature $\tilde{\bm{z}}_{t\rightarrow 0}^{i-1,s\times warp \times \frac{1}{s}}$. The $Conv(\cdot)$ operation denotes a zero-initialized convolutional layer, which prevents the intermediate features in $DU(\cdot)$ from being overly influenced during the early training stages. Furthermore, in the $t$-th diffusion step, after we obtain the denoised features $\bm{z}_{t-1}^{i}$, we can predict the noise-free approximation $\tilde{\bm{z}}_{t-1\rightarrow 0}^{i}$ of the current feature through the reverse process described in Eq. \eqref{con:z_0}. This noise-free approximation will be used as the guiding information for the adjacent frames in the next diffusion step. After $T$ steps, the final features $\{\bm{z}_{1}^{i}\}_{i=1}^{N}$ are fed into a pretrained VAE decoder $\mathcal{D}(\cdot)$, yielding the high-resolution output frames $\{\bm{y}^{i}\}_{i=1}^{N}$.

%% file: sec/5_experiment.tex
\begin{figure*}
  \includegraphics[width=\textwidth]{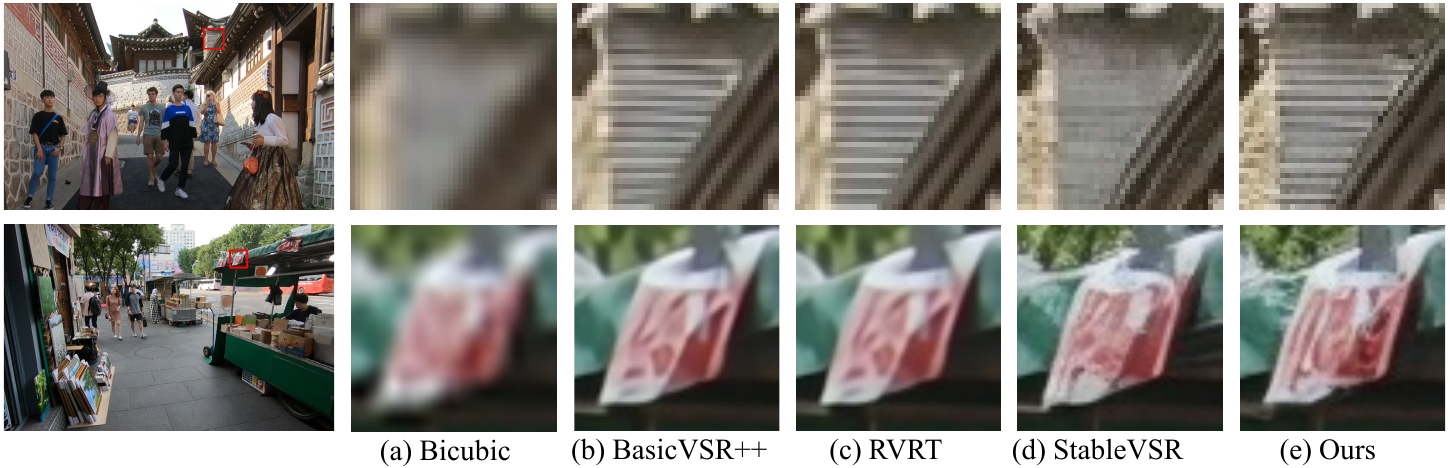}
  \vspace{-1.5em}
  \caption{Visualization of the VSR results from different methods on REDS4 dataset. }
  \label{fig:x4}
\end{figure*}

\section{Experiments}
\label{Experiments}

\subsection{Experimental Settings}\label{Experimental Settings}
The descriptions of the experimental settings for our DGAF-VSR are provided in Supplementary 8.
Discussion with video-based pretrained DM is provided in Supplementary 9.

\subsection{Comparison with State-of-the-art Methods}
We evaluate the effectiveness of DGAF-VSR in comparison to eight state-of-the-art (SOTA) VSR methods, including four non-DM-based methods EDVR \cite{wang2019edvr}, BasicVSR \cite{chan2021basicvsr}, BasicVSR++ \cite{chan2022basicvsr++}, and RVRT \cite{liang2022rvrt}, as well as six DM-based methods StableVSR \cite{rota2024StableVSR}, MGLD-VSR \cite{yang2024MGLD}, UAV \cite{zhou2023upscaleavideo}, STAR \cite{xie2025star}, DOVE \cite{chen2025dove} and SeedVR2 \cite{wang2025seedvr2}. The quantitative and qualitative comparisons are carried out on the synthetic datasets REDS \cite{Nah_2019_CVPR_Workshops_REDS} and Vid4 \cite{liu2013bayesian}, as well as the real-world dataset VideoLQ \cite{chan2022investigating}.

\textbf{Quantitative Evaluation.} 
The quantitative results on the synthetic datasets REDS4 and Vid4 are shown in Table \ref{tab:x4}. As can be seen, our DGAF-VSR achieves the best performance on all perceptual metrics, including both full-reference metrics (LPIPS, DISTS) and no-reference metrics (MUSIQ, CLIP-IQA, NIQE). Specifically, compared to the top-performing non-DM-based method, DGAF-VSR enhances the LPIPS metric by 25.78\% on the REDS4 dataset. When compared to the SOTA DM-based method, DGAF-VSR achieves an improvement of 0.75 in MUSIQ on the Vid4 dataset. The impressive perceptual performance of DGAF-VSR reflects its strong capability to reconstruct realistic details and textures for the VSR task.

In addition, compared to DM-based methods, our DGAF-VSR achieves the highest PSNR and SSIM scores on synthetic datasets, demonstrating a superior balance between fidelity and perceptual quality. Specifically, DGAF-VSR outperforms StableVSR by 0.20 dB in PSNR and 0.009 in SSIM for the REDS4 dataset, and by 0.28 dB in PSNR and 0.015 in SSIM for the Vid4 dataset. Furthermore, our DGAF-VSR demonstrates excellent VSR performance on the real-world VideoLQ dataset, as shown in Table \ref{tab:jvxu_q3}. Specifically, DGAF-VSR outperforms STAR by 4.46 in MUSIQ value and 59.11\% in CLIP-IQA value. These quantitative results confirm the superior reconstruction ability of our proposed network in VSR. Comparisons of VSR performance at different scale factors are presented in Supplementary 10.

\begin{table}[htbp]
\caption{Quantitative comparison on the VideoLQ for $4\times$ VSR in terms of three perceptual metrics, with the best results in bold.}
\label{tab:jvxu_q3}
\begin{center}
\centering
\renewcommand{\arraystretch}{0.85}
\small
\begin{tabular}{c|ccc}
\toprule[1.0pt]
\rowcolor{mygray} Method & MUSIQ$\uparrow$ &CLIP-IQA$\uparrow$ & DOVER$\uparrow$  \\ \hline\hline
UAV \cite{zhou2023upscaleavideo}& 30.88  & 0.182  &  0.4493 \\
MGLD-VSR \cite{yang2024MGLD}& 55.99  & 0.339  &  0.7254 \\
STAR \cite{xie2025star} & 54.59  &  0.313 & 0.7034   \\
Ours  & \textbf{59.05}  &  \textbf{0.498}  &  \textbf{0.7599}  \\
\toprule[1.0pt]
\end{tabular}
\end{center}
\end{table}

\textbf{Qualitative Evaluation.} 
Figure \ref{fig:x4} visualizes the $4\times$ VSR results of different methods on the REDS4 dataset. As can be seen, our method effectively maintains high-frequency information while recovering realistic image details, such as the texture of the wall and the patterns on the food trolley signs. In contrast, other approaches suffer from blurred edges \cite{chan2022basicvsr++,liang2022rvrt} or unrealistic details \cite{rota2024StableVSR}. Additional comparisons on the Vid4 dataset are provided in the Supplementary 11. 
Descriptions of the high-frequency preservation capabilities of DGAF-VSR are shown in Supplementary 12.

\subsection{Temporal Consistency of DGAF-VSR}
\begin{figure*}
  \includegraphics[width=\textwidth]{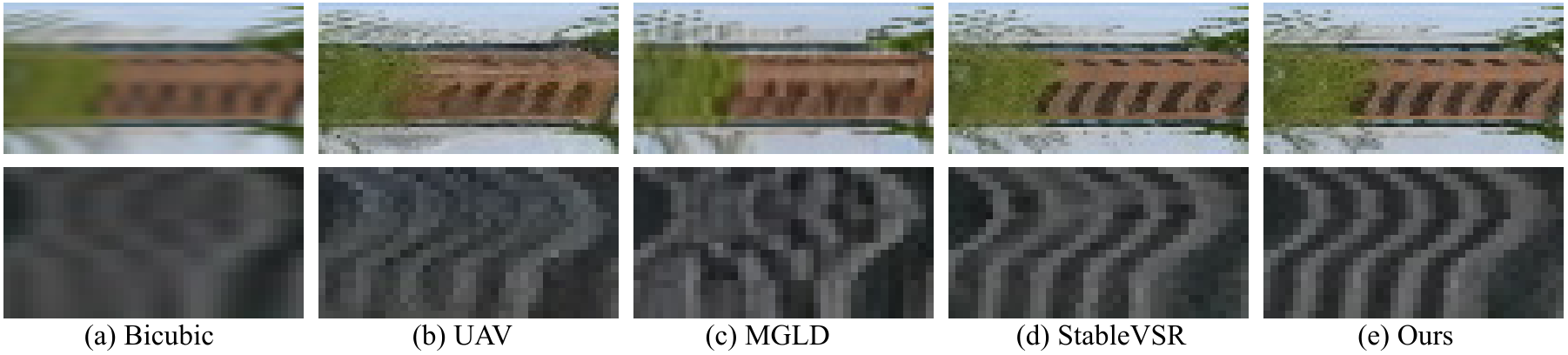}
  \vspace{-1.5em}
  \caption{Visualization of the temporal profiles from different methods on REDS4 dataset. We examine a row and track changes across adjacent frames over time.}
  \vspace{-1em}
  \label{fig:tc_vis}
\end{figure*}
\begin{table*}[htbp]
\caption{The effect of the dense guidance mechanism and the rescaling-based warping strategy, in terms of two perceptual ($\diamond$), two fidelity ($\star$) and two temporal consistency ($\circ$) metrics, with the best results in bold and second bests underlined.}
\label{tab:ablation}
\begin{center}
\renewcommand{\arraystretch}{0.85}
\setlength{\tabcolsep}{0.8mm}
\begin{tabular}{c|cc|c|cccccc}
\toprule
\rowcolor{mygray} 
Case & Bicubic & Nearest & Dense guidance & PSNR$\star\uparrow$ & SSIM$\star\uparrow$ & LPIPS$\diamond\downarrow$ & DISTS$\diamond\downarrow$ & tLPIPS$\circ\downarrow$ & tOF$\circ\downarrow$ \\ \hline \midrule 
1 &  & \checkmark &  & 27.81 & 0.791 & 0.104 & 0.047 & 6.25 & 2.83 \\ 
2 &  &  & \checkmark & 26.70 & 0.756 & 0.118 & 0.050 & 21.61 & 3.29 \\
3 & \checkmark &  & \checkmark & \underline{28.13} & \underline{0.803} & \underline{0.099} & \underline{0.045} & \underline{4.28} & \underline{2.81} \\
4(Ours) &  & \checkmark & \checkmark & \textbf{28.17}&	\textbf{0.804}&	\textbf{0.095}&	\textbf{0.043}&	\textbf{3.92}	&\textbf{2.71} \\ 
\bottomrule
\end{tabular}
\vspace{-1em}
\end{center}
\end{table*}
\begin{figure}
\centering
\includegraphics[width=0.4\textwidth]{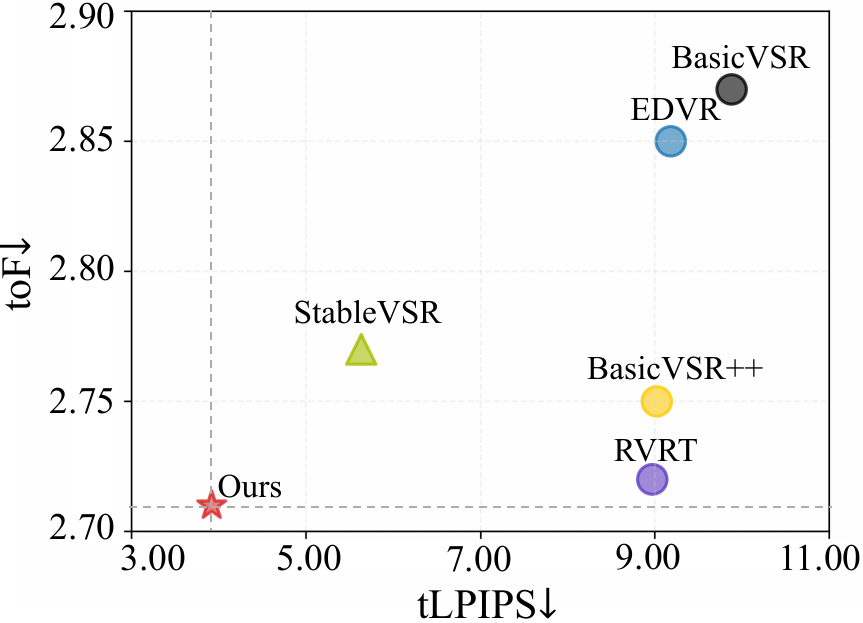}
\caption{The temporal consistency results for VSR on REDS4 dataset. }
\label{fig:tc}
\end{figure}

According to Observation 1, our approach leverages strong inter- and intra-variable correlations to achieve superior temporal consistency in VSR results. In this section, we systematically validate this advantage.
As shown in Figure \ref{fig:tc}, we evaluate the temporal consistency metrics tLPIPS \cite{Chu_2020} and tOF \cite{Chu_2020} for the VSR output sequences on the REDS4 dataset across different approaches. Specifically, our approach shows a 56.30\% improvement in tLPIPS over RVRT, a SOTA non-DM-based method. In addition, compared to the SOTA DM-based method StableVSR, our approach improves tLPIPS by 30.37\% and tOF by 2.02\%. 
Furthermore, in Figure \ref{fig:tc_vis}, we show a comparison of the temporal profiles between our DGAF-VSR and SOTA DM-based methods UAV, MGLD and StableVSR. As illustrated, the profiles from UAV and MGLD exhibit noticeable noise, indicating flickering artifacts. The profile from StableVSR still shows visible discontinuities, while the profile of our approach exhibits a more seamless and smoother transition. These quantitative gains and qualitative observations collectively confirm the effectiveness of our feature alignment and dense guidance mechanism in enhancing temporal consistency.

\subsection{Efficiency of DGAF-VSR}
The running efficiency of the network is a crucial factor to consider for practical applications. We present the runtime, network parameters, and the GPU memory overhead of DGAF-VSR in Supplementary 13.

\subsection{Ablation Study}
In this subsection, we analyze the effects of several crucial components on the proposed DGAF-VSR, including the rescaling-based warping strategy in the OGWM modules, the dense guidance mechanism in the FTCM modules, the in-pair forward and backward guidance strategy, the pre-upscaling strategy, the choice of optical flow estimator and the rescaling factor. All ablation studies are conducted on the REDS dataset, and the analyzes of the last four ablation studies are provided in the Supplementary 14.

\textbf{Effect of the dense guidance mechanism.} 
As shown in Table \ref{tab:ablation}, a comparison between case 1 and case 4 evaluates the $4\times$ VSR performance with and without the dense guidance mechanism. Incorporating this strategy improves PSNR by 0.36 dB, lowers the LPIPS score by 0.009, and reduces tLPIPS score by 37.28\%, indicating significant improvements in fidelity, perceptual quality, and temporal consistency.
In conclusion, these results substantiate the effectiveness of the dense guidance mechanism.

\textbf{Effect of the rescaling-based warping strategy.} 
The case 2, case 3 and case 4 in Table \ref{tab:ablation} compare the effects of bicubic and nearest neighbor interpolation operations. Specifically, in case 2, when the warping process is applied to the original features without any interpolation, the PSNR value is 26.70 dB, with an SSIM of 0.756, an LPIPS of 0.118, and a tOF of 3.29. In case 3, employing the bicubic interpolation method improves the VSR performance, yielding a PSNR of 28.13 dB, an SSIM of 0.803, and decreases the LPIPS and tOF to 0.099 and 2.81, respectively. Furthermore, our proposed method using nearest neighbor interpolation, as shown in case 4, achieves the highest PSNR of 28.17 dB, an SSIM value of 0.804, and the lowest LPIPS and tOF, which are 0.095 and 2.71 respectively. These results suggest that performing warping at a higher resolution significantly enhances VSR performance, while nearest neighbor interpolation outperforms bicubic interpolation by better preserving spatial details and temporal coherence.

%% file: sec/6_conclusion.tex
\section{Conclusion}
In this paper, we propose DGAF-VSR, a novel diffusion model (DM)-based network for video super-resolution (VSR).
Our method significantly enhances perceptual quality and reconstruction fidelity while maintaining robust temporal consistency.
Within the DM-based pipeline, we identify two key observations: (a) the feature domain exhibits stronger spatial and temporal correlations than the pixel domain, and (b) warping at an upscaled resolution is highly effective for preserving high-frequency information during alignment, but this improvement is not necessarily monotonic.
Leveraging these two observations, for each diffusion step within the DM-based pipeline, we develop an optical guided warping module (OGWM) for precise feature alignments, as well as a feature-wise temporal condition module (FTCM) for dense guidance from adjacent features. 
Experimental results demonstrated that our DGAF-VSR achieved state-of-the-art VSR performance in terms of perceptual quality, fidelity, and temporal consistency. Exploring more effective feature warping and dense guidance mechanisms across a broader range of video restoration tasks presents a promising direction for future research.

%% file: main.bbl
\begin{thebibliography}{64}
\providecommand{\natexlab}[1]{#1}
\providecommand{\url}[1]{\texttt{#1}}
\expandafter\ifx\csname urlstyle\endcsname\relax
  \providecommand{\doi}[1]{doi: #1}\else
  \providecommand{\doi}{doi: \begingroup \urlstyle{rm}\Url}\fi

\bibitem[Baker and Kanade(1999)]{baker1999super}
Simon Baker and Takeo Kanade.
\newblock \emph{Super-resolution optical flow}, volume 510.
\newblock Carnegie Mellon University, The Robotics Institute Pittsburgh, 1999.

\bibitem[Fransens et~al.(2007)Fransens, Strecha, and Van~Gool]{fransens2007optical}
Rik Fransens, Christoph Strecha, and Luc Van~Gool.
\newblock Optical flow based super-resolution: A probabilistic approach.
\newblock \emph{Computer vision and image understanding}, 106\penalty0 (1):\penalty0 106--115, 2007.

\bibitem[Brox et~al.(2004)Brox, Bruhn, Papenberg, and Weickert]{brox2004high}
Thomas Brox, Andr{\'e}s Bruhn, Nils Papenberg, and Joachim Weickert.
\newblock High accuracy optical flow estimation based on a theory for warping.
\newblock In \emph{Computer Vision-ECCV 2004: 8th European Conference on Computer Vision, Prague, Czech Republic, May 11-14, 2004. Proceedings, Part IV 8}, pages 25--36. Springer, 2004.

\bibitem[Liu and Sun(2013)]{liu2013bayesian}
Ce~Liu and Deqing Sun.
\newblock On bayesian adaptive video super resolution.
\newblock \emph{IEEE transactions on pattern analysis and machine intelligence}, 36\penalty0 (2):\penalty0 346--360, 2013.

\bibitem[Chan et~al.(2021)Chan, Wang, Yu, Dong, and Loy]{chan2021basicvsr}
Kelvin~CK Chan, Xintao Wang, Ke~Yu, Chao Dong, and Chen~Change Loy.
\newblock {BasicVSR}: The search for essential components in video super-resolution and beyond.
\newblock In \emph{Proceedings of the IEEE/CVF conference on computer vision and pattern recognition}, pages 4947--4956, 2021.

\bibitem[Liang et~al.(2022)Liang, Fan, Xiang, Ranjan, Ilg, Green, Cao, Zhang, Timofte, and Gool]{liang2022rvrt}
Jingyun Liang, Yuchen Fan, Xiaoyu Xiang, Rakesh Ranjan, Eddy Ilg, Simon Green, Jiezhang Cao, Kai Zhang, Radu Timofte, and Luc~V Gool.
\newblock Recurrent video restoration transformer with guided deformable attention.
\newblock \emph{Advances in Neural Information Processing Systems}, 35:\penalty0 378--393, 2022.

\bibitem[Tian et~al.(2020)Tian, Zhang, Fu, and Xu]{tian2020tdan}
Yapeng Tian, Yulun Zhang, Yun Fu, and Chenliang Xu.
\newblock {TDAN}: Temporally-deformable alignment network for video super-resolution.
\newblock In \emph{Proceedings of the IEEE/CVF conference on computer vision and pattern recognition}, pages 3360--3369, 2020.

\bibitem[Wang et~al.(2019)Wang, Chan, Yu, Dong, and Change~Loy]{wang2019edvr}
Xintao Wang, Kelvin~CK Chan, Ke~Yu, Chao Dong, and Chen Change~Loy.
\newblock {EDVR}: Video restoration with enhanced deformable convolutional networks.
\newblock In \emph{Proceedings of the IEEE/CVF conference on computer vision and pattern recognition workshops}, pages 0--0, 2019.

\bibitem[Chan et~al.(2022{\natexlab{a}})Chan, Zhou, Xu, and Loy]{chan2022basicvsr++}
Kelvin~CK Chan, Shangchen Zhou, Xiangyu Xu, and Chen~Change Loy.
\newblock Basicvsr++: Improving video super-resolution with enhanced propagation and alignment.
\newblock In \emph{Proceedings of the IEEE/CVF conference on computer vision and pattern recognition}, pages 5972--5981, 2022{\natexlab{a}}.

\bibitem[Chan et~al.(2022{\natexlab{b}})Chan, Zhou, Xu, and Loy]{chan2022investigating}
Kelvin~CK Chan, Shangchen Zhou, Xiangyu Xu, and Chen~Change Loy.
\newblock Investigating tradeoffs in real-world video super-resolution.
\newblock In \emph{Proceedings of the IEEE/CVF Conference on Computer Vision and Pattern Recognition}, pages 5962--5971, 2022{\natexlab{b}}.

\bibitem[Zhou et~al.(2024{\natexlab{a}})Zhou, Zhang, Zhao, Wang, Li, and Gu]{zhou2024video}
Xingyu Zhou, Leheng Zhang, Xiaorui Zhao, Keze Wang, Leida Li, and Shuhang Gu.
\newblock Video super-resolution transformer with masked inter\&intra-frame attention.
\newblock In \emph{Proceedings of the IEEE/CVF Conference on Computer Vision and Pattern Recognition}, pages 25399--25408, 2024{\natexlab{a}}.

\bibitem[Zhou et~al.(2024{\natexlab{b}})Zhou, Yang, Wang, Luo, and Loy]{zhou2023upscaleavideo}
Shangchen Zhou, Peiqing Yang, Jianyi Wang, Yihang Luo, and Chen~Change Loy.
\newblock {Upscale-A-Video}: Temporal-consistent diffusion model for real-world video super-resolution.
\newblock In \emph{Proceedings of the IEEE/CVF Conference on Computer Vision and Pattern Recognition}, pages 2535--2545, 2024{\natexlab{b}}.

\bibitem[Rota et~al.(2024)Rota, Buzzelli, and van~de Weijer]{rota2024StableVSR}
Claudio Rota, Marco Buzzelli, and Joost van~de Weijer.
\newblock Enhancing perceptual quality in video super-resolution through temporally-consistent detail synthesis using diffusion models.
\newblock In \emph{European Conference on Computer Vision}, pages 36--53. Springer, 2024.

\bibitem[Chen et~al.(2024)Chen, Long, Qiu, Yao, Zhou, Luo, and Mei]{chen2024SATeCo}
Zhikai Chen, Fuchen Long, Zhaofan Qiu, Ting Yao, Wengang Zhou, Jiebo Luo, and Tao Mei.
\newblock Learning spatial adaptation and temporal coherence in diffusion models for video super-resolution.
\newblock In \emph{Proceedings of the IEEE/CVF Conference on Computer Vision and Pattern Recognition}, pages 9232--9241, 2024.

\bibitem[Xue et~al.(2019)Xue, Chen, Wu, Wei, and Freeman]{xue2019TOFlow}
Tianfan Xue, Baian Chen, Jiajun Wu, Donglai Wei, and William~T Freeman.
\newblock Video enhancement with task-oriented flow.
\newblock \emph{International Journal of Computer Vision}, 127:\penalty0 1106--1125, 2019.

\bibitem[Dai et~al.(2017)Dai, Qi, Xiong, Li, Zhang, Hu, and Wei]{dai2017DCN}
Jifeng Dai, Haozhi Qi, Yuwen Xiong, Yi~Li, Guodong Zhang, Han Hu, and Yichen Wei.
\newblock Deformable convolutional networks.
\newblock In \emph{Proceedings of the IEEE international conference on computer vision}, pages 764--773, 2017.

\bibitem[Cao et~al.(2021)Cao, Li, Zhang, and Van~Gool]{cao2021vsrtransformer}
Jiezhang Cao, Yawei Li, Kai Zhang, and Luc Van~Gool.
\newblock Video super-resolution transformer.
\newblock \emph{arXiv preprint arXiv:2106.06847}, 2021.

\bibitem[Liang et~al.(2024)Liang, Cao, Fan, Zhang, Ranjan, Li, Timofte, and Van~Gool]{liang2024vrt}
Jingyun Liang, Jiezhang Cao, Yuchen Fan, Kai Zhang, Rakesh Ranjan, Yawei Li, Radu Timofte, and Luc Van~Gool.
\newblock {VRT}: A video restoration transformer.
\newblock \emph{IEEE Transactions on Image Processing}, 2024.

\bibitem[Shi et~al.(2022)Shi, Gu, Xie, Wang, Yang, and Dong]{shi2022PSRT}
Shuwei Shi, Jinjin Gu, Liangbin Xie, Xintao Wang, Yujiu Yang, and Chao Dong.
\newblock Rethinking alignment in video super-resolution transformers.
\newblock \emph{Advances in Neural Information Processing Systems}, 35:\penalty0 36081--36093, 2022.

\bibitem[Xu et~al.(2024{\natexlab{a}})Xu, Park, Zhang, Zhou, Shechtman, Liu, Huang, and Liu]{xu2024videogigagan}
Yiran Xu, Taesung Park, Richard Zhang, Yang Zhou, Eli Shechtman, Feng Liu, Jia-Bin Huang, and Difan Liu.
\newblock Videogigagan: Towards detail-rich video super-resolution.
\newblock \emph{arXiv preprint arXiv:2404.12388}, 2024{\natexlab{a}}.

\bibitem[Kang et~al.(2023)Kang, Zhu, Zhang, Park, Shechtman, Paris, and Park]{kang2023gigagan}
Minguk Kang, Jun-Yan Zhu, Richard Zhang, Jaesik Park, Eli Shechtman, Sylvain Paris, and Taesung Park.
\newblock Scaling up gans for text-to-image synthesis.
\newblock In \emph{Proceedings of the IEEE/CVF conference on computer vision and pattern recognition}, pages 10124--10134, 2023.

\bibitem[Saharia et~al.(2022)Saharia, Ho, Chan, Salimans, Fleet, and Norouzi]{saharia2021SR3}
Chitwan Saharia, Jonathan Ho, William Chan, Tim Salimans, David~J Fleet, and Mohammad Norouzi.
\newblock Image super-resolution via iterative refinement.
\newblock \emph{IEEE transactions on pattern analysis and machine intelligence}, 45\penalty0 (4):\penalty0 4713--4726, 2022.

\bibitem[Wang et~al.(2024)Wang, Yue, Zhou, Chan, and Loy]{wang2024StableSR}
Jianyi Wang, Zongsheng Yue, Shangchen Zhou, Kelvin~CK Chan, and Chen~Change Loy.
\newblock Exploiting diffusion prior for real-world image super-resolution.
\newblock \emph{International Journal of Computer Vision}, 132\penalty0 (12):\penalty0 5929--5949, 2024.

\bibitem[Yue et~al.(2024)Yue, Liao, and Loy]{yue2025InvSR}
Zongsheng Yue, Kang Liao, and Chen~Change Loy.
\newblock Arbitrary-steps image super-resolution via diffusion inversion.
\newblock \emph{arXiv preprint arXiv:2412.09013}, 2024.

\bibitem[Yue et~al.(2023)Yue, Wang, and Loy]{ResShift}
Zongsheng Yue, Jianyi Wang, and Chen~Change Loy.
\newblock {ResShift}: Efficient diffusion model for image super-resolution by residual shifting.
\newblock \emph{Advances in Neural Information Processing Systems}, 36:\penalty0 13294--13307, 2023.

\bibitem[Wu et~al.(2024)Wu, Sun, Ma, and Zhang]{wu2024OSEDiff}
Rongyuan Wu, Lingchen Sun, Zhiyuan Ma, and Lei Zhang.
\newblock One-step effective diffusion network for real-world image super-resolution.
\newblock \emph{Advances in Neural Information Processing Systems}, 37:\penalty0 92529--92553, 2024.

\bibitem[Yeh et~al.(2024)Yeh, Lin, Wang, Hsiao, Chen, Shiu, and Liu]{yeh2024diffir2vr}
Chang-Han Yeh, Chin-Yang Lin, Zhixiang Wang, Chi-Wei Hsiao, Ting-Hsuan Chen, Hau-Shiang Shiu, and Yu-Lun Liu.
\newblock Diffir2vr-zero: Zero-shot video restoration with diffusion-based image restoration models.
\newblock \emph{arXiv preprint arXiv:2407.01519}, 2024.

\bibitem[Yang et~al.(2024)Yang, He, Ma, and Zhang]{yang2024MGLD}
Xi~Yang, Chenhang He, Jianqi Ma, and Lei Zhang.
\newblock Motion-guided latent diffusion for temporally consistent real-world video super-resolution.
\newblock In \emph{European Conference on Computer Vision}, pages 224--242. Springer, 2024.

\bibitem[Huang et~al.(2017)Huang, Wang, and Wang]{7919264}
Yan Huang, Wei Wang, and Liang Wang.
\newblock Video super-resolution via bidirectional recurrent convolutional networks.
\newblock \emph{IEEE transactions on pattern analysis and machine intelligence}, 40\penalty0 (4):\penalty0 1015--1028, 2017.

\bibitem[Fuoli et~al.(2019)Fuoli, Gu, and Timofte]{fuoli2019efficientvideosuperresolutionrecurrent}
Dario Fuoli, Shuhang Gu, and Radu Timofte.
\newblock Efficient video super-resolution through recurrent latent space propagation.
\newblock In \emph{2019 IEEE/CVF International Conference on Computer Vision Workshop (ICCVW)}, pages 3476--3485. IEEE, 2019.

\bibitem[Jo et~al.(2018)Jo, Oh, Kang, and Kim]{Jo_2018_DUF}
Younghyun Jo, Seoung~Wug Oh, Jaeyeon Kang, and Seon~Joo Kim.
\newblock Deep video super-resolution network using dynamic upsampling filters without explicit motion compensation.
\newblock In \emph{Proceedings of the IEEE conference on computer vision and pattern recognition}, pages 3224--3232, 2018.

\bibitem[Xu et~al.(2024{\natexlab{b}})Xu, Yu, Wang, Mi, and Yao]{xu2024IART}
Kai Xu, Ziwei Yu, Xin Wang, Michael~Bi Mi, and Angela Yao.
\newblock Enhancing video super-resolution via implicit resampling-based alignment.
\newblock In \emph{Proceedings of the IEEE/CVF Conference on Computer Vision and Pattern Recognition}, pages 2546--2555, 2024{\natexlab{b}}.

\bibitem[Ho et~al.(2020)Ho, Jain, and Abbeel]{ho2020DDPM}
Jonathan Ho, Ajay Jain, and Pieter Abbeel.
\newblock Denoising diffusion probabilistic models.
\newblock \emph{arXiv preprint arxiv:2006.11239}, 2020.

\bibitem[Rombach et~al.(2022{\natexlab{a}})Rombach, Blattmann, Lorenz, Esser, and Ommer]{rombach2021LDM}
Robin Rombach, Andreas Blattmann, Dominik Lorenz, Patrick Esser, and Bj{\"o}rn Ommer.
\newblock High-resolution image synthesis with latent diffusion models.
\newblock In \emph{Proceedings of the IEEE/CVF conference on computer vision and pattern recognition}, pages 10684--10695, 2022{\natexlab{a}}.

\bibitem[Nah et~al.(2019)Nah, Baik, Hong, Moon, Son, Timofte, and Mu~Lee]{Nah_2019_CVPR_Workshops_REDS}
Seungjun Nah, Sungyong Baik, Seokil Hong, Gyeongsik Moon, Sanghyun Son, Radu Timofte, and Kyoung Mu~Lee.
\newblock Ntire 2019 challenge on video deblurring and super-resolution: Dataset and study.
\newblock In \emph{Proceedings of the IEEE/CVF conference on computer vision and pattern recognition workshops}, pages 0--0, 2019.

\bibitem[Wang et~al.(2004)Wang, Bovik, Sheikh, and Simoncelli]{wang2004image}
Zhou Wang, Alan~C Bovik, Hamid~R Sheikh, and Eero~P Simoncelli.
\newblock Image quality assessment: from error visibility to structural similarity.
\newblock \emph{IEEE transactions on image processing}, 13\penalty0 (4):\penalty0 600--612, 2004.

\bibitem[Hore and Ziou(2010)]{hore2010image}
Alain Hore and Djemel Ziou.
\newblock Image quality metrics: {PSNR} vs. {SSIM}.
\newblock In \emph{2010 20th international conference on pattern recognition}, pages 2366--2369. IEEE, 2010.

\bibitem[Shannon(1948)]{shannon1948mathematical}
Claude~E Shannon.
\newblock A mathematical theory of communication.
\newblock \emph{The Bell system technical journal}, 27\penalty0 (3):\penalty0 379--423, 1948.

\bibitem[Altman and Bland(2005)]{altman2005standard}
Douglas~G Altman and J~Martin Bland.
\newblock Standard deviations and standard errors.
\newblock \emph{Bmj}, 331\penalty0 (7521):\penalty0 903, 2005.

\bibitem[Zhou et~al.(2018)Zhou, Li, and Li]{zhou2018high}
Fuqiang Zhou, Xiaojie Li, and Zuoxin Li.
\newblock High-frequency details enhancing densenet for super-resolution.
\newblock \emph{Neurocomputing}, 290:\penalty0 34--42, 2018.

\bibitem[Li et~al.(2023)Li, Zhang, Liu, and Zhu]{li2023feature}
Ao~Li, Le~Zhang, Yun Liu, and Ce~Zhu.
\newblock Feature modulation transformer: Cross-refinement of global representation via high-frequency prior for image super-resolution.
\newblock In \emph{Proceedings of the IEEE/CVF International Conference on Computer Vision}, pages 12514--12524, 2023.

\bibitem[Fritsche et~al.(2019)Fritsche, Gu, and Timofte]{fritsche2019frequency}
Manuel Fritsche, Shuhang Gu, and Radu Timofte.
\newblock Frequency separation for real-world super-resolution.
\newblock In \emph{2019 IEEE/CVF International Conference on Computer Vision Workshop (ICCVW)}, pages 3599--3608. IEEE, 2019.

\bibitem[Canny(1986)]{canny1986computational}
John Canny.
\newblock A computational approach to edge detection.
\newblock \emph{IEEE Transactions on pattern analysis and machine intelligence}, \penalty0 (6):\penalty0 679--698, 1986.

\bibitem[Sobel et~al.(1968)Sobel, Feldman, et~al.]{sobel19683x3}
Irwin Sobel, Gary Feldman, et~al.
\newblock A 3x3 isotropic gradient operator for image processing.
\newblock \emph{a talk at the Stanford Artificial Project in}, 1968:\penalty0 271--272, 1968.

\bibitem[Jain et~al.(1995)Jain, Kasturi, Schunck, et~al.]{jain1995machine}
Ramesh Jain, Rangachar Kasturi, Brian~G Schunck, et~al.
\newblock \emph{Machine vision}, volume~5.
\newblock McGraw-hill New York, 1995.

\bibitem[Van~der Schaaf and van Hateren(1996)]{van1996modelling}
van~A Van~der Schaaf and JH~van van Hateren.
\newblock Modelling the power spectra of natural images: statistics and information.
\newblock \emph{Vision research}, 36\penalty0 (17):\penalty0 2759--2770, 1996.

\bibitem[Nussbaumer and Nussbaumer(1982)]{nussbaumer1982fast}
Henri~J Nussbaumer and Henri~J Nussbaumer.
\newblock \emph{The fast Fourier transform}.
\newblock Springer, 1982.

\bibitem[Shepard(1968)]{shepard1968two}
Donald Shepard.
\newblock A two-dimensional interpolation function for irregularly-spaced data.
\newblock In \emph{Proceedings of the 1968 23rd ACM national conference}, pages 517--524, 1968.

\bibitem[Teed and Deng(2020)]{teed2020raft}
Zachary Teed and Jia Deng.
\newblock {RAFT}: Recurrent all-pairs field transforms for optical flow.
\newblock In \emph{Computer Vision--ECCV 2020: 16th European Conference, Glasgow, UK, August 23--28, 2020, Proceedings, Part II 16}, pages 402--419. Springer, 2020.

\bibitem[He et~al.(2024)He, Xue, Liu, Lin, Gao, Lin, Qiao, Ouyang, and Liu]{he2024venhancer}
Jingwen He, Tianfan Xue, Dongyang Liu, Xinqi Lin, Peng Gao, Dahua Lin, Yu~Qiao, Wanli Ouyang, and Ziwei Liu.
\newblock {VEnhancer}: Generative space-time enhancement for video generation.
\newblock \emph{arXiv preprint arXiv:2407.07667}, 2024.

\bibitem[Xie et~al.(2025)Xie, Liu, Zhou, Zhao, Zhou, Zhang, Zhang, Yang, Yang, and Tai]{xie2025star}
Rui Xie, Yinhong Liu, Penghao Zhou, Chen Zhao, Jun Zhou, Kai Zhang, Zhenyu Zhang, Jian Yang, Zhenheng Yang, and Ying Tai.
\newblock Star: Spatial-temporal augmentation with text-to-video models for real-world video super-resolution.
\newblock \emph{arXiv preprint arXiv:2501.02976}, 2025.

\bibitem[Chu et~al.(2020)Chu, Xie, Mayer, Leal-Taix{\'e}, and Thuerey]{Chu_2020}
Mengyu Chu, You Xie, Jonas Mayer, Laura Leal-Taix{\'e}, and Nils Thuerey.
\newblock Learning temporal coherence via self-supervision for gan-based video generation.
\newblock \emph{ACM Transactions on Graphics (TOG)}, 39\penalty0 (4):\penalty0 75--1, 2020.

\bibitem[Rombach et~al.(2022{\natexlab{b}})Rombach, Blattmann, Lorenz, Esser, and Ommer]{Rombach_2022_CVPR}
Robin Rombach, Andreas Blattmann, Dominik Lorenz, Patrick Esser, and Bj{\"o}rn Ommer.
\newblock High-resolution image synthesis with latent diffusion models.
\newblock In \emph{Proceedings of the IEEE/CVF conference on computer vision and pattern recognition}, pages 10684--10695, 2022{\natexlab{b}}.

\bibitem[{Stability AI}(2022)]{sdx4upscaler}
{Stability AI}.
\newblock {stabilityai/stable-diffusion-x4-upscaler}.
\newblock \url{https://huggingface.co/stabilityai/stable-diffusion-x4-upscaler}, 2022.

\bibitem[Blattmann et~al.(2023)Blattmann, Dockhorn, Kulal, Mendelevitch, Kilian, Lorenz, Levi, English, Voleti, Letts, et~al.]{blattmann2023stablevideodiffusionscaling}
Andreas Blattmann, Tim Dockhorn, Sumith Kulal, Daniel Mendelevitch, Maciej Kilian, Dominik Lorenz, Yam Levi, Zion English, Vikram Voleti, Adam Letts, et~al.
\newblock Stable video diffusion: Scaling latent video diffusion models to large datasets.
\newblock \emph{arXiv preprint arXiv:2311.15127}, 2023.

\bibitem[Zhang et~al.(2023)Zhang, Rao, and Agrawala]{zhang2023ControlNet}
Lvmin Zhang, Anyi Rao, and Maneesh Agrawala.
\newblock Adding conditional control to text-to-image diffusion models.
\newblock In \emph{Proceedings of the IEEE/CVF international conference on computer vision}, pages 3836--3847, 2023.

\bibitem[Zhang et~al.(2018)Zhang, Isola, Efros, Shechtman, and Wang]{zhang2018unreasonableeffectivenessdeepfeatures}
Richard Zhang, Phillip Isola, Alexei~A Efros, Eli Shechtman, and Oliver Wang.
\newblock The unreasonable effectiveness of deep features as a perceptual metric.
\newblock In \emph{Proceedings of the IEEE conference on computer vision and pattern recognition}, pages 586--595, 2018.

\bibitem[Ding et~al.(2020)Ding, Ma, Wang, and Simoncelli]{Ding_2020}
Keyan Ding, Kede Ma, Shiqi Wang, and Eero~P Simoncelli.
\newblock Image quality assessment: Unifying structure and texture similarity.
\newblock \emph{IEEE transactions on pattern analysis and machine intelligence}, 44\penalty0 (5):\penalty0 2567--2581, 2020.

\bibitem[Ke et~al.(2021)Ke, Wang, Wang, Milanfar, and Yang]{ke2021musiqmultiscaleimagequality}
Junjie Ke, Qifei Wang, Yilin Wang, Peyman Milanfar, and Feng Yang.
\newblock {MUSIQ}: Multi-scale image quality transformer.
\newblock In \emph{Proceedings of the IEEE/CVF international conference on computer vision}, pages 5148--5157, 2021.

\bibitem[Wang et~al.(2023)Wang, Chan, and Loy]{wang2022exploringclipassessinglook}
Jianyi Wang, Kelvin~CK Chan, and Chen~Change Loy.
\newblock Exploring clip for assessing the look and feel of images.
\newblock In \emph{Proceedings of the AAAI conference on artificial intelligence}, volume~37, pages 2555--2563, 2023.

\bibitem[Mittal et~al.(2012)Mittal, Soundararajan, and Bovik]{6353522}
Anish Mittal, Rajiv Soundararajan, and Alan~C Bovik.
\newblock Making a “completely blind” image quality analyzer.
\newblock \emph{IEEE Signal processing letters}, 20\penalty0 (3):\penalty0 209--212, 2012.

\bibitem[Li et~al.(2025)Li, Liu, Cao, Chen, Zhuang, Chen, He, Wang, and Qiao]{li2025diffvsr}
Xiaohui Li, Yihao Liu, Shuo Cao, Ziyan Chen, Shaobin Zhuang, Xiangyu Chen, Yinan He, Yi~Wang, and Yu~Qiao.
\newblock Diffvsr: Enhancing real-world video super-resolution with diffusion models for advanced visual quality and temporal consistency.
\newblock \emph{arXiv e-prints}, pages arXiv--2501, 2025.

\bibitem[Podell et~al.(2023)Podell, English, Lacey, Blattmann, Dockhorn, M{\"u}ller, Penna, and Rombach]{podell2023sdxl}
Dustin Podell, Zion English, Kyle Lacey, Andreas Blattmann, Tim Dockhorn, Jonas M{\"u}ller, Joe Penna, and Robin Rombach.
\newblock {SDXL}: Improving latent diffusion models for high-resolution image synthesis.
\newblock \emph{arXiv preprint arXiv:2307.01952}, 2023.

\bibitem[Esser et~al.(2024)Esser, Kulal, Blattmann, Entezari, M{\"u}ller, Saini, Levi, Lorenz, Sauer, Boesel, et~al.]{esser2024scaling}
Patrick Esser, Sumith Kulal, Andreas Blattmann, Rahim Entezari, Jonas M{\"u}ller, Harry Saini, Yam Levi, Dominik Lorenz, Axel Sauer, Frederic Boesel, et~al.
\newblock Scaling rectified flow transformers for high-resolution image synthesis.
\newblock In \emph{Forty-first international conference on machine learning}, 2024.

\end{thebibliography}
